\documentclass[twocolumn,10pt]{article}
\usepackage[top=1in, bottom=1in, left=1in, right=1in]{geometry}
\usepackage{amsmath,amsfonts,amscd,amssymb,graphicx}

\newcommand{\bb}{\mathbf{b}}

\newcommand{\bd}{\mathbf{d}}

\newcommand{\bx}{\mathbf{x}}

\newcommand{\bA}{\mathbf{A}}

\newcommand{\bW}{\mathbf{W}}
\newcommand{\bX}{\mathbf{X}}

\newcommand{\bAtilde}{\mathbf{\tilde{A}}}

\newcommand{\bVtilde}{\mathbf{\tilde{V}}}
\newcommand{\bSigmatilde}{\boldsymbol{\tilde{\Sigma}}}

\newcommand{\bphi}{\boldsymbol{\phi}}
\newcommand{\bPhi}{\boldsymbol{\Phi}}

\newcommand{\bLambda}{\boldsymbol{\Lambda}}
\newcommand{\bOmega}{\boldsymbol{\Omega}}

\DeclareMathOperator*{\argmin}{arg\rm{}min}

\usepackage{dcolumn}
\usepackage{bm}
\usepackage{overpic}
\usepackage{color}
\usepackage{adjustbox}
\usepackage[caption=false]{subfig}
\usepackage{rotating}
\usepackage{float}

\usepackage{algorithm}
\usepackage{algpseudocode}
\algnewcommand\algorithmicinput{\textbf{Input:}}
\algnewcommand\Input{\item[\algorithmicinput]}

\begin{document}

\title{Bagging, optimized dynamic mode decomposition (BOP-DMD) for robust, stable forecasting with spatial and temporal uncertainty-quantification}

\author{Diya Sashidhar and J. Nathan Kutz\\
Department of Applied Mathematics, University of Washington, Seattle, WA 98195-3925} 

 
\maketitle
 
\begin{abstract}
{\em Dynamic mode decomposition} (DMD) provides a regression framework for adaptively learning a best-fit linear dynamics model over snapshots of temporal, or spatio-temporal, data.  A diversity of regression techniques have been developed for producing the linear model approximation whose solutions are exponentials in time.  For spatio-temporal data, DMD provides low-rank and interpretable models in the form of dominant modal structures along with their exponential/oscillatory behavior in time.
The majority of DMD algorithms, however, are prone to bias errors from noisy measurements of the dynamics, leading to poor model fits and unstable forecasting capabilities. The optimized DMD algorithm minimizes the model bias with a variable projection optimization, thus leading to stabilized forecasting capabilities. Here, the optimized DMD algorithm is improved by using statistical bagging methods whereby a single set of snapshots is used to produce an {\em ensemble} of optimized DMD models. The outputs of these models are averaged to produce a {\em bagging, optimized dynamic mode decomposition} (BOP-DMD).  BOP-DMD not only improves performance, it also robustifies the model and provides both spatial and temporal uncertainty quantification (UQ).  Thus unlike currently available DMD algorithms, BOP-DMD provides a stable and robust model for {\em probabilistic}, or Bayesian forecasting with comprehensive UQ metrics.
\end{abstract}

\section{INTRODUCTION} \label{intro}

\begin{figure}[t]
    \centering
    \begin{overpic}[width = .45\textwidth]{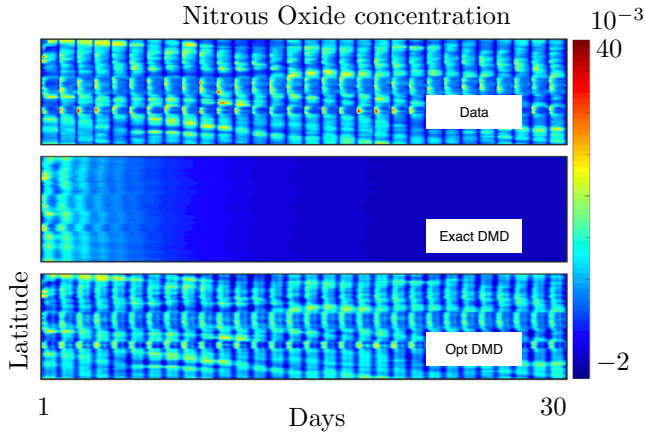}
    \put(26,72){Nitrous Oxide concentration}
    \put(-5,10){\rotatebox{90}{Latitude}}
    \put(45,0){Days}
    \put(0,2){1}
    \put(90,2){30}
    \put(100,66){$40$}
    \put(100,10){$-2$}
    \put(99,71){$10^{-3}$}
    \end{overpic}
    \caption{Canonical bias effect in DMD algorithms for noisy (normalized) data (top panel) for a specific longitude and elevation.  In this example of atmospheric chemistry from Velagar et al~\cite{velegar2019scalable}, thirty days of nitrous oxide (NO) are measured and a DMD model regression is used to fit the data which has been normalized~\cite{velegar2019scalable}.  The bias of the exact DMD algorithm (middle panel) shows that the solution almost immediately tends to zero while optimized DMD (bottom panel) is able to correctly approximate the chemical dynamics.}
    \label{fig:atmos_chem}
\end{figure}

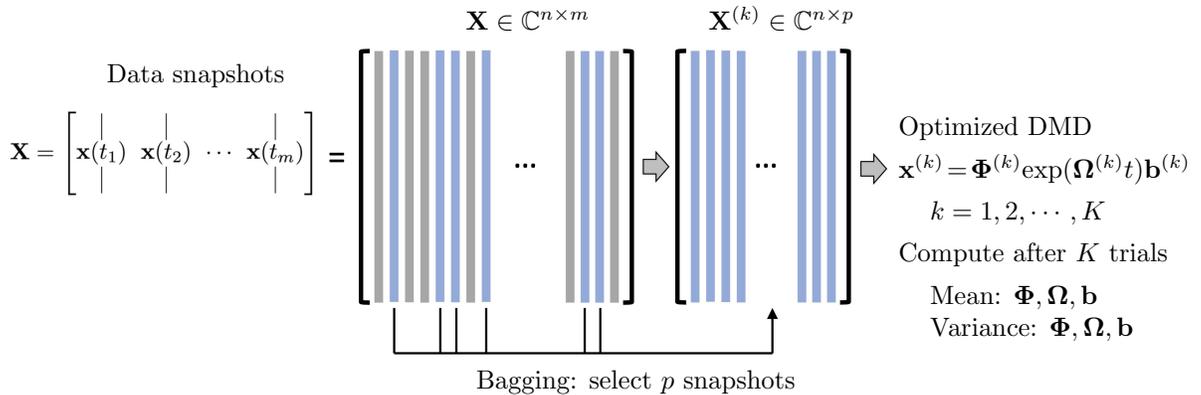
\begin{figure*}[t]
\centering
\begin{overpic}[width=.85\textwidth]{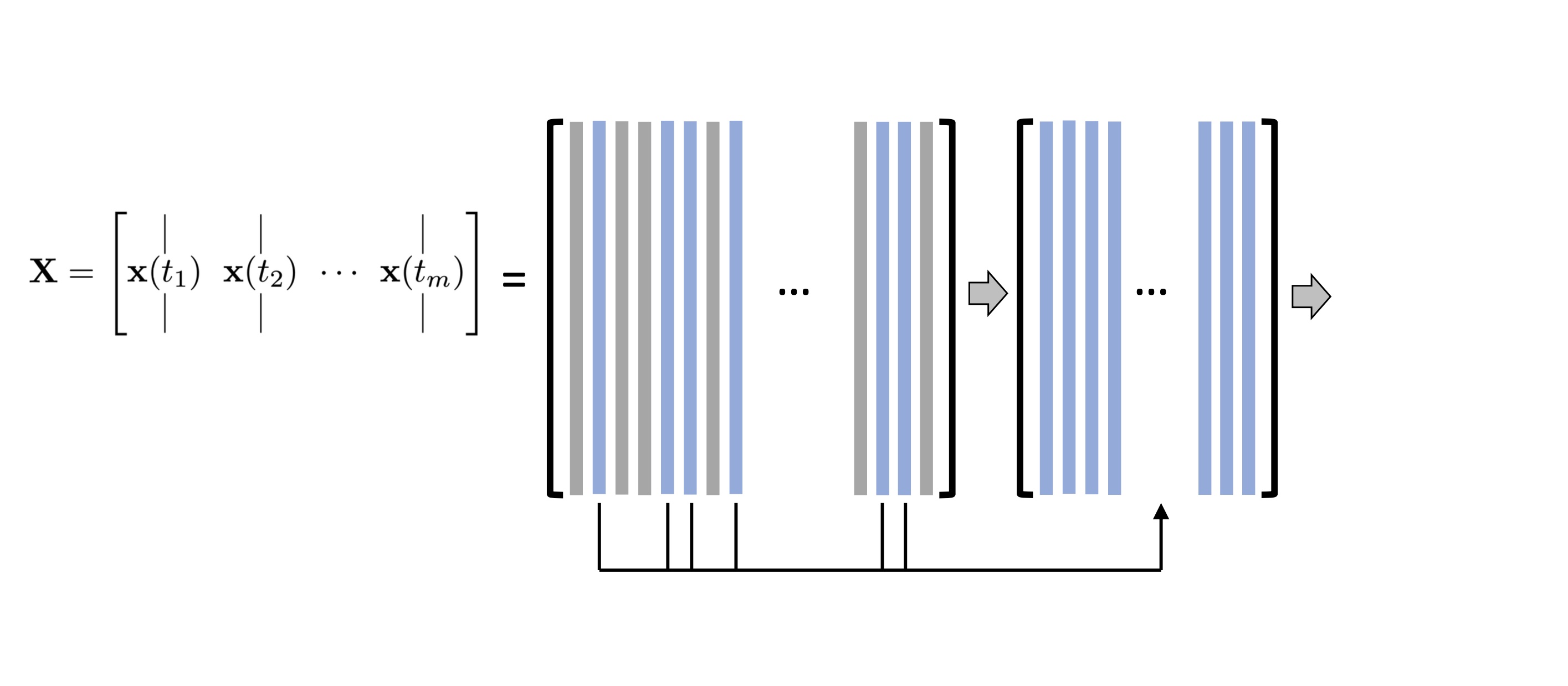}
\put(86,24){$\bx^{(k)}\!=\! \bPhi^{(k)} \! \exp(\bOmega^{(k)} t)\bb^{(k)}$}
\put(45,38){$\bX\in\mathbb{C}^{n\times m}$}
\put(68,38){$\bX^{(k)}\in\mathbb{C}^{n\times p}$}
\put(89,20){$k=1,2, \cdots , K$}
\put(46,4){Bagging: select $p$ snapshots}
\put(86,16){Compute after $K$ trials}
\put(86,28){Optimized DMD}
\put(89,12){Mean: $\bPhi, \bOmega, \bb$ }
\put(89,9){Variance: $\bPhi, \bOmega, \bb$}
\put(11,33){Data snapshots}
\end{overpic}
\vspace*{-.2in}
\caption{
\label{fig:BOPDMD}
Summary of the BOP-DMD architecture.  The data snapshots $\bx(t_k)$ are collected over $m$ snapshots into the matrix $\bX$.  Columns of $\bX$ are randomly sub-selected into the matrix $\bX^{(k)}$ to build an optimized DMD model.  Each DMD model $\bx^{(k)}\!=\! \bPhi^{(k)} \! \exp(\bOmega^{(k)} t)\bb^{(k)}$ is used to compute the statistics (mean and variance) of the DMD parametrizations $\bPhi, \bOmega, \bb$ which are used in building a the BOP-DMD ensemble solution with UQ. }
\end{figure*}

The data-driven modeling and control of complex systems, which includes the ability to produce accurate and robust forecasting algorithms, is a rapidly evolving field with the potential to transform the engineering, biological, and physical sciences.  The combination of high-fidelity measurements from modern sensor technologies and numerical simulations has ensured that while data is often abundant, accurate and computationally efficient models for forecasting remain challenging. {\em Dynamic mode decomposition} (DMD) provides a data-driven regression architecture for adaptively learning linear dynamics models over snapshots of temporal data.  Although often used for the discovery of approximate dynamical systems from high-dimensional, spatio-temporal data (e.g. fluid flows), it can equally applied to simple time series measurements to produce best-fit models.  DMD has been broadly used in the scientific community due to its ease of use, interpretability and adaptive nature~\cite{kutz2016book}. However, bias induced by measurement noise has significantly limited the forecasting and reconstruction performance of DMD algorithms, as illustrated in the example measurements of atmospheric chemistry dynamics of Fig.~\ref{fig:atmos_chem}.  As shown in Fig.~\ref{fig:BOPDMD}, using an optimized DMD formulation, we show that leveraging the statistical method of bagging ({\bf b}ootstrap {\bf agg}regat{\bf ing})~\cite{breiman1984classification} significantly improves DMD robustness and  accuracy for forecasting while also providing {\em uncertainty quantification} (UQ) in its predictions and reconstructions.  

Dynamic mode decomposition originated in the fluid dynamics community.  Introduced as as algorithm by Schmid~\cite{schmid2008aps,schmid2010jfm}, it has rapidly become a commonly used data-driven analysis tool and the standard algorithm to approximate the Koopman operator from data~\cite{rowley2009jfm}.  In the context of fluid dynamics, DMD was used to identify spatio-temporal coherent fluid structures from high-dimensional time-series data.  The DMD analysis offered an alternative to standard dimensionality reduction methods such as the {\em proper orthogonal decomposition} (POD), which highlighted low-rank features in fluid flows using the computationally efficient {\em singular value decomposition} (SVD)~\cite{kutz:2013}.  The advantage of using DMD over SVD is that the DMD modes are linear combinations of the SVD modes that have a common linear (exponential) behavior in time, given by oscillations at a fixed frequency with growth or decay.  Specifically, DMD is a regression to solutions of the form
\begin{align}
\bx(t) = \sum_{j=1}^r \bphi_j e^{\omega_j t} b_j = \bPhi \exp(\bOmega t)\bb,
\label{eq:DMDapprox}
\end{align}
where $\bx(t)$ is an $r$-rank approximation to a collection of state space measurements $\mathbf{x}_k=\mathbf{x}(t_k)$ $(k=1, 2, \cdots , n)$.  The algorithm regresses to values of the DMD eigenvalues $\omega_j$, DMD modes $\bphi_j$ and their loadings $b_j$.  The $\omega_j$ determines the temporal behavior of the system associated with a modal structure $\bphi_j$, thus giving a highly interpretable representation of the dynamics.  Such a regression can also be learned from time-series data~\cite{lange2020fourier}.  DMD may be thought of as a combination of SVD/POD in space with the Fourier transform in time, combining the strengths of each approach~\cite{chen2012jns,kutz2016book}.  DMD is modular due to its simple formulation in terms of linear algebra, resulting in innovations related to control~\cite{proctor2016siads}, compression~\cite{brunton2015jcd,erichson2016arxiva}, reduced-order modeling~\cite{alla2017nonlinear}, and multi-resolution analysis~\cite{kutz2016multiresolution,champion2019discovery}, among others.

Because of its simplicity and interpretability, DMD has been applied to a wide range of diverse applications beyond fluid mechanics, including neuroscience~\cite{brunton2016extracting}, disease modeling~\cite{proctor2015ih}, robotics~\cite{mamakoukas2019proc,mamakoukas2020arxiv}, video processing~\cite{grosek2014arxiv,erichson2016jrtp}, power grids~\cite{susuki2009,susuki2011nonlinear}, financial markets~\cite{mann2016qf}, and plasma physics~\cite{taylor2018dynamic,kaptanoglu2020pop}.  The regression to (\ref{eq:DMDapprox}) shows the immediate value of DMD for forecasting.  Specifically, any time $t^*$ can be evaluated to produce an approximation to the state of the system $\bx(t^*)$.  However, despite its introduction more than a decade ago, DMD is rarely used for forecasting and/or reconstruction of time-series data except in cases with high-quality (noise-free or nearly noise free) data.  Indeed, practitioners who work with DMD and noisy data know that the algorithm fails not only to produce a reasonable forecast, but also often fails in the task of reconstructing the time-series it was originally regressed to.  Thus in the past decade, the value of DMD has largely been as an important {\em diagnostic} tool as the DMD modes and frequencies are highly interpretable.  Indeed, from  Schmid's~\cite{schmid2008aps,schmid2010jfm} original work until now, DMD papers are primarily diagnostic in nature with the key figures of any given paper being the DMD modes and eigenvalues.  In cases, where DMD is used on noise-free data, such as for producing reduced order models from high-fidelity numerical simulation data~\cite{bagheri2014,alla2017nonlinear}, then the DMD solution~(\ref{eq:DMDapprox}) can be used for reconstructing and forecasting accurate representations of the solution.

The origins of the DMD's poor performance for reconstruction and forecasting is due to bias-induced effects from noisy measurements~\cite{bagheri2014,hemati2017biasing,dawson2016ef,askham2017arxiv}.  Indeed, based on observations of the DMD spectrum from Bagheri et al~\cite{bagheri2014} and Degennaro et al~\cite{degennaro2015uncertainty},
Hemati et al.~\cite{hemati2017biasing} identified the explicit origins of this noise-driven bias effect on the eigevalue distribution and proposed methods that could circumvent it.  Following~\cite{hemati2017biasing,dawson2016ef}, Askham and Kutz~\cite{askham2017arxiv} produced the {\em optimized} DMD (optDMD) algorithm which is a nonlinear optimization enabled by variable projection techniques which provides optimal de-biasing for a given signal-to-noise ratio. Indeed, optDMD is the standard for producing the most accurate and precise DMD modes and eigenvalues in the presence of noise.  However, the variable projection method often fails to converge, limiting its usefulness.
In what follows, we empower the optimized DMD algorithm with three critical new features:  (i) an initialization procedure allowing for stabilizing the convergence of the variable projection algorithm, (ii) a statistical bagging scheme~\cite{breiman1984classification}, whose objective is to reduce variance and stabilize models, for improving the accuracy and robustness of the regression, and (iii) uncertainty quantification (UQ) in the DMD model, both temporal and spatial.   Our algorithm, termed the {\em bagging, optimized dynamic mode decomposition} (BOP-DMD) is
highlighted in Fig.~\ref{fig:BOPDMD} and
is characterized against current DMD algorithms, demonstrating its  performance for not only diagnostics, but also reconstruction and probabilistic forecasting. It is further applied to a number of example spatio-temporal systems.
It should be noted that alternatives to improve DMD models include choosing appropriate windows of time for sampling  data~\cite{kutz2016multiresolution,heydari2021modal} and ensembling DMD regressions on short burst of data, which was developed by Scandurra, Tezzele and Louiseau for the pyDMD package (Tutorial 8)~\cite{demo18pydmd}.  These sampling strategies can also be used to great effect with the BOP-DMD algorithm, with the ensembling method being similar in spirit, but formalized and improved here with bagging and optimization.
BOP-DMD also easily integrates other methods for improving DMD models, including centering data~\cite{hirsh2019} and time-delay embedding data~\cite{brunton2017natcomm,arbabi2017,kamb2020siads,hirsh2021structured}.  Importantly, BOP-DMD can be used on top of these algorithms instead of the exact DMD algorithm for improving accuracy, stability and performance while generating UQ metrics, reducing variance and preventing over-fitting through bagging~\cite{breiman1984classification}.

\section{ALGORITHMS FOR DMD}    

The algorithmic construction of the DMD method can be best understood from the so-called {\em exact} DMD~\cite{tu2014jcd}.  Indeed, this exact DMD is simply a least-square fitting procedure.  Specifically, the DMD algorithm seeks a best fit linear operator $\mathbf{A}$ that approximately advances the state of a system, $\mathbf{x}\in\mathbb{R}^n$, forward in time according to the linear dynamical system
\begin{align}\label{Eq:DMD:Propagator}
    \mathbf{x}_{k+1} = \mathbf{A}\mathbf{x}_k,
\end{align}
where $\mathbf{x}_k=\mathbf{x}(k\Delta t)$, and $\Delta t$ denotes a fixed time step that is small enough to resolve the highest frequencies in the dynamics.  Thus, the operator $\mathbf{A}$ is an approximation of the Koopman operator $\mathcal{K}$ restricted to a measurement subspace spanned by direct measurements of the state $\mathbf{x}$~\cite{rowley2009jfm}.
%
%
In the original DMD formulation~\cite{schmid2010jfm}, uniform sampling in time was required so that 
$t_k=k\Delta t$.  The exact DMD algorithm~\cite{tu2014jcd} does not require uniform sampling.  Rather, for each snapshot $\mathbf{x}(t_k)$ there is a corresponding snapshot $\mathbf{x}(t_k')$ one time step $\Delta t$ in the future.  These snapshots are arranged into two matrices, $\bX$ and $\bX'$:
\begin{subequations}
\begin{align}
\bX &= \begin{bmatrix} \vline & \vline & & \vline \\
\bx(t_1) & \bx(t_2) & \cdots & \bx(t_m) \\
 \vline & \vline & & \vline
 \end{bmatrix}\\
 \bX' &= \begin{bmatrix} \vline & \vline & & \vline \\
\bx(t_1') & \bx(t_2') & \cdots & \bx(t_m') \\
 \vline & \vline & & \vline
 \end{bmatrix}.
\end{align}
\end{subequations}
In terms of these matrices, the DMD regression~\eqref{Eq:DMD:Propagator} is
\begin{align}
\bX' \approx \bA \bX. \label{eq:DMD-matrix}
\end{align}
The exact DMD is the best fit, in a least-squares sense, operator $\bA$ that approximately advances snapshot measurements forward in time.  Specifically, it can be formulated as an optimization problem
\begin{align}
\bA = \argmin_{\bA} \|\bX' - \bA \bX\|_F = \bX'\bX^\dagger\label{Eq:DMD:Definition}
\end{align}
where $\|\cdot\|_F$ is the Frobenius norm and $^\dagger$ denotes the pseudo-inverse.  The pseudo-inverse may be computed using the SVD of $\mathbf{X}=\mathbf{U}\boldsymbol{\Sigma}\mathbf{V}^*$ as $\mathbf{X}^\dagger=\mathbf{V}\boldsymbol{\Sigma}^{-1}\mathbf{U}^*$.  The matrices $\mathbf{U}\in\mathbb{C}^{n\times n}$ and $\mathbf{V}^{m\times m}$ are unitary, so that $\mathbf{U}^*\mathbf{U}=\mathbf{I}$ and $\mathbf{V}^*\mathbf{V}=\mathbf{I}$, where $^*$ denotes complex-conjugate transpose.  The columns of $\mathbf{U}$ are known as POD modes.
Often for high-dimensional data, the DMD leverages low-rank structure
by first projecting $\mathbf{A}$ onto the first $r$ POD modes in $\mathbf{U}_r$ and approximating the pseudo-inverse using the rank-$r$ SVD approximation $\bX\approx \mathbf{U}_r\boldsymbol{\Sigma}_r\mathbf{V}_r^*$:
\begin{align}
\tilde{\mathbf{A}} = \mathbf{U}_r^*\mathbf{A}\mathbf{U}_r  = \mathbf{U}_r^*\mathbf{X}'\mathbf{V}_r\boldsymbol{\Sigma}_r^{-1}.
\end{align}

The leading spectral decomposition of $\mathbf{A}$ may be approximated from the spectral decomposition of the much smaller $\bAtilde$:
\begin{align}
\bAtilde \bW = \bW \bLambda.
\end{align}
The diagonal matrix $\bLambda$ contains the \emph{DMD eigenvalues}, which correspond to eigenvalues of the high-dimensional matrix $\bA$.
The columns of $\bW$ are eigenvectors of $\bAtilde$, and provide a coordinate transformation that diagonalizes the matrix.  These columns may be thought of as linear combinations of POD mode amplitudes that behave linearly with a single temporal pattern given by the corresponding eigenvalue $\lambda$.

The eigenvectors of $\bA$ are the \emph{DMD modes} $\bPhi$, and they are reconstructed using the eigenvectors $\bW$ of the reduced system and the time-shifted data matrix $\bX'$:
\begin{align}
\bPhi = \bX' \bVtilde \bSigmatilde^{-1}\bW.
\end{align}
Tu et al.~\cite{tu2014jcd} proved that these DMD modes are eigenvectors of the full $\bA$ matrix under certain conditions.  As already shown in the introduction, the DMD decomposition allows for a reconstruction of the solution as (\ref{eq:DMDapprox}).  The amplitudes of each mode $\bb$ can be computed from
\begin{align}
\bb = \bPhi^{\dagger}\bx_1.\label{Eq:DMDModeAmplitude},
\end{align}
however, alternative and often better approaches are available~$\bb$~\cite{chen2012jns,jovanovic2014pof,askham2017arxiv}.
Thus, the data matrix $\mathbf{X}$ may be reconstructed as
\begin{eqnarray}
      \bX &\approx&  \bPhi \mbox{diag}(\bb) {\bf T}(\boldsymbol{\omega})
   \\
  =&& \!\!\!\! \!\!\!\! \!\!\!\!  \!\!\!\!\!\! \left[ \! \begin{array}{ccc} | & & | \\ \boldsymbol{\phi}_1 & 
    \!\!\cdots\!\! & \boldsymbol{\phi}_r \\ | & & | \end{array} \! \right] \!\!
    \left[ \! \begin{array}{ccc} b_1 &  & \\ & \!\!\ddots\!\! & \\ & & b_r  \end{array} \!\right] \!\!
    \left[ \! \begin{array}{ccc} e^{\omega_1 t_1} & \!\cdots\! & e^{\omega_1 t_m} \\
    \vdots & \!\ddots\! & \vdots \\ e^{\omega_r t_1} & \!\cdots\! & e^{\omega_r t_m} \end{array} \!\! \right]
    . \nonumber
\label{eq:dmd_opt}
\end{eqnarray}
Bagheri~\cite{bagheri2014} first highlighted that DMD is particularly sensitive to the effects of noisy data, with systematic biases introduced to the eigenvalue distribution~\cite{duke2012error,bagheri2013jfm,dawson2016ef,hemati2017tcfd}.  As a result, a number of methods have been introduced to stabilize performance, including total least-squares DMD~\cite{hemati2017tcfd}, forward-backward DMD~\cite{dawson2016ef}, variational DMD~\cite{azencot2019consistent}, subspace DMD~\cite{takeishi2017}, time-delay embedded DMD~\cite{brunton2017natcomm,arbabi2017,kamb2020siads,hirsh2021structured} and robust DMD methods~\cite{askham2017arxiv,scherl2020prf}.
However, the {\em optimized DMD} algorithm of Askham and Kutz~\cite{askham2017arxiv}, which uses a variable projection method for nonlinear least squares to compute the DMD for unevenly timed samples, provides the best and optimal performance of any algorithm currently available.  This is not surprising given that it actually is constructed to optimally satisfy the DMD problem formulation.  Specifically, the optimized DMD algorithm solves the exponential fitting problem directly:
\begin{equation}
       \argmin_{ \boldsymbol{\omega}, \bPhi_{\bf b} } \| \bX -   \bPhi_{\bf b} {\bf T}(\boldsymbol{\omega}) \|_F,
\end{equation}
where $\bPhi_b$ = $\bPhi \bf{diag(b)}$. This has been shown to provide a superior decomposition due to its ability to optimally suppress bias and handle snapshots collected at arbitrary times.  The disadvantage of optimized DMD is that one must solve a nonlinear optimization problem, often which can fail to converge.

\section{BOP-DMD:  BAGGING, OPTIMIZED DMD}

BOP-DMD leverages Breiman's statistical bagging sampling strategy~\cite{breiman1984classification} shown in Fig.~\ref{fig:BOPDMD} along with the optimized DMD algorithm.  Bagging is designed to produce an ensemble of models, thereby reducing model variance and suppressing over-fitting by design.  Not only is ensembling improve DMD, it also is effective in deep neural network regressions~\cite{allen2020towards}.  Further innovations here include stabilizing the variable projection technique used by optDMD so that it converges consistently to an optimal solution. 
Its ability to converge is often dependent upon a suitable initial guess for the DMD eigenvalues and eigenvectors. More precisely, we have observed through simulations that the initial guess of the eigenvectors has the most pronounced effect on the reliable convergence of variable projection.  In its standard form, no initial guess is given and failure to converge is common for complex, noisy data.

\begin{algorithm}[t]
		\caption{BOP-DMD}
		\label{alg:bopdmd}
		\begin{algorithmic}
			\Input Input ($\bX$, $p$, $K$)  
			\Procedure{BOPDMD}{${\bf X}$, $p$, $K$}
			\State Compute $\bPhi_0, \bOmega_0, \bb_0$  
			\Comment{optDMD regression}
	\For{$k \in \{ 1, 2, \cdots, K \}$}
	\Comment{Compute $K$ optDMD models.}	
	        \State Choose $p$ of $m$ snapshots ($p<m$)
	        \Comment Bagging	
			\State  opt-DMD $\bPhi_k, \bOmega_k, \bb_k$
			\Comment{Initialize with $\bOmega_0$}
			\State Update $\bPhi_0, \bOmega_0, \bb_0$
			\Comment{Use improved $\bPhi, \bOmega, \bb$}
			\EndFor			
			\State Compute mean $\boldsymbol{\mu}=\{ \langle \bPhi \rangle, \langle\bOmega \rangle, \langle\bb \rangle \}$
			\State Compute variance $\boldsymbol{\sigma}=\{ \langle\bPhi^2 \rangle,\langle \bOmega^2 \rangle,\langle \bb^2 \rangle \}$			
			\State \textbf{return} $\boldsymbol{\mu}, \boldsymbol{\sigma}$
			\Comment{Return optDMD parameters.}
			\EndProcedure
		\end{algorithmic}
\end{algorithm}

The BOP-DMD algorithm accounts for the initialization process and further provides the optimal solutions to linear models by using opt-DMD as the regression architecture.  Algorithm~\ref{alg:bopdmd} shows the algorithmic structure of BOP-DMD, highlighting the bagging, initialization and ensembling of the DMD models to produce an ensemble, probabilistic DMD model.  The initialization of DMD is accomplished by first constructing an optDMD model approximation, whose eigenvalues and eigenvectors $\bPhi_0$ can be used to seed the BOP-DMD.

Algorithm~\ref{alg:bopdmd} provides a robust method for producing DMD models.  As will be shown in the next section, BOP-DMD outperforms the opt-DMD algorithm which has already been shown to be the gold standard for producing DMD models~\cite{askham2017arxiv,azencot2019consistent}.  In addition to the DMD model itself, the output of the algorithm can be used to produce a forecast with UQ metrics as shown in Algorithm~\ref{alg:forecast}.  Specifically, the BOP-DMD model is used in a Monte-Carlo fashion to produce a probabilistic forecast for a given system.  This forecasting is accomplished by simply drawing from the probability distributions of the $\bPhi, \bOmega, \bb$ computed from BOP-DMD.  The mean and variance of the forecast are produced, giving uncertainty quantification with the forecasting prediction.  Both Algorithms~\ref{alg:bopdmd} and \ref{alg:forecast} are demonstrated in what follows.

\begin{algorithm}[t]
		\caption{Forecasting BOP-DMD}
		\label{alg:forecast}
		\begin{algorithmic}
			\Input Input ($\boldsymbol{\mu}, \boldsymbol{\sigma}, {\bf T}, K$ )  
			\Procedure{forecastBOPDMD}{$\boldsymbol{\mu}, \boldsymbol{\sigma},{\bf T},K$}
%
%
	\For{$k \in \{ 1, 2, \cdots, K \}$}
	\Comment{Compute $K$ Forecasts.}
	        \State Generate $\langle \bPhi\rangle, \bOmega_k, \bb_k$
	        \Comment Drawn from ($\boldsymbol{\mu}, \boldsymbol{\sigma}$) 
			\State  ${\bx}_k ({\bf T}) =  \langle \bPhi\rangle \exp(\bOmega_k {\bf T}) \bb_k$ 
			\Comment  Solution for times ${\bf   T}$
			\EndFor
			\State Compute mean $ \langle \bx ({\bf T}) \rangle $
			\State Compute variance $ \langle \bx^2 ({\bf T}) \rangle $
			\State \textbf{return} $ \langle \bx ({\bf T}) \rangle $, $ \langle \bx^2 ({\bf T}) \rangle $
			\Comment{Return optDMD forecast }
			\EndProcedure
		\end{algorithmic}
\end{algorithm}


\begin{figure}[t]
    \centering
    \begin{overpic}[width = .45\textwidth]{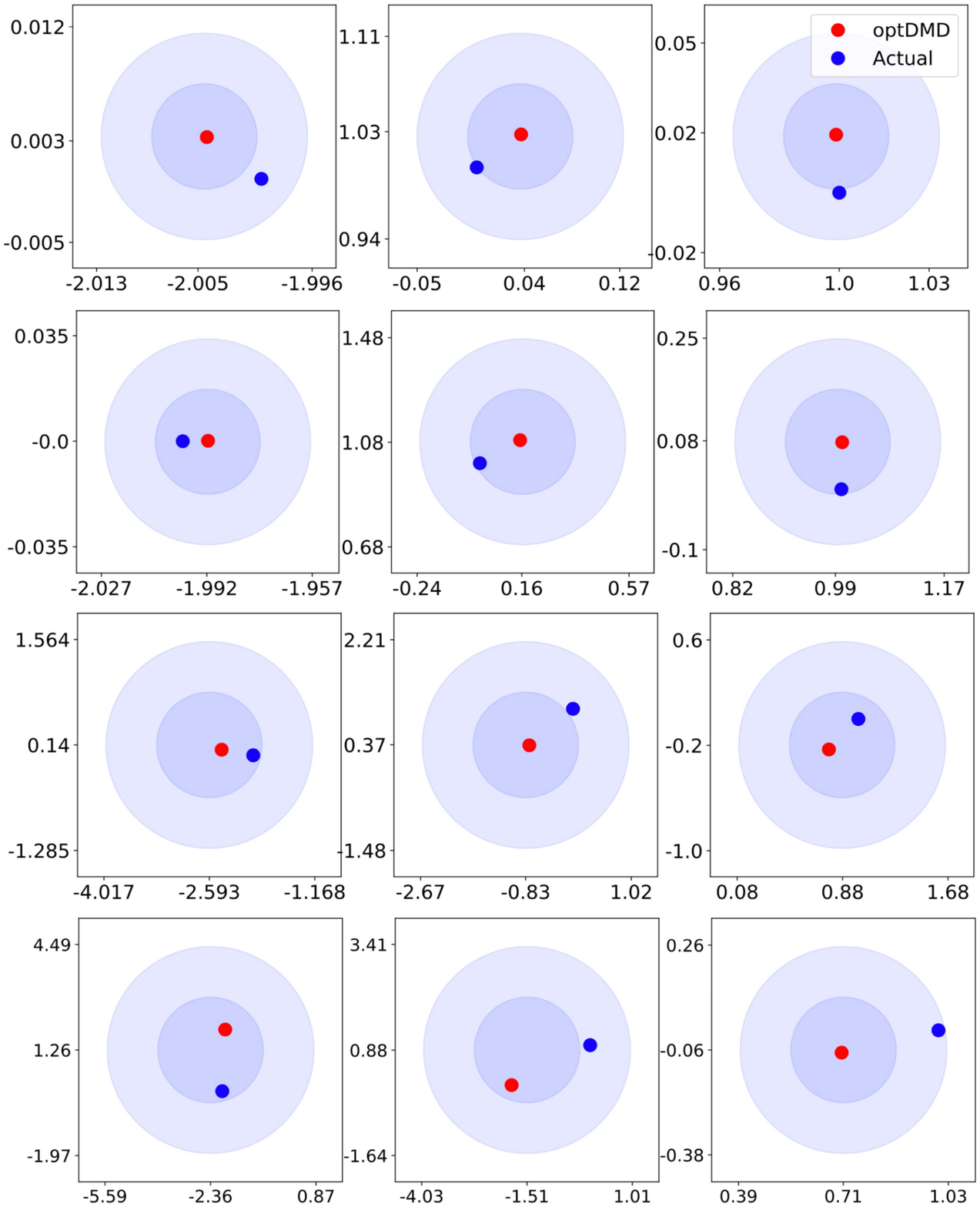}
    \put(-1,20){(d)}
     \put(-1,43){(c)}
      \put(-1,66){(b)}
       \put(-1,90){(a)}
    \put(16,97){$\lambda_1$}
    \put(40,97){$\lambda_2$}
    \put(63,97){$\lambda_3$}
    \put(14,1){$\Re\{\lambda\}$}
    \put(0,12){\rotatebox{90}{$\Im\{\lambda\}$}}
    \put(8,93){low noise}
    \put(8,69){medium noise}
    \put(8,46){high noise}
    \put(8,23){corruption}
    \end{overpic}
    \caption{Uncertainty quantification for computed eigenvalue magnitudes for 4 separate noise realizations (a - d) for eigenvalues $\lambda_1 = -2$, $\lambda_2  = i$, $\lambda_3 = 1$. For comparison, the magnitude of the true eigenvalue is depicted by a solid blue dot, the magnitude of the computed eigenvalue using optDMD is depicted by a solid red dot. The two concentric circles represent the range of eigenvalue magnitudes within one and two standard deviations, respectively. These results were generating using  k = 100 trials of the BOP-DMD algorithm with a randomly selected subset of size p = 20 for low and medium noise settings (a and b) and p = 50 for higher noise settings (c and d).}
    \label{fig:noise_cycles}
\end{figure}

\begin{figure}[t]
    \centering
    \includegraphics[width = .53\textwidth]{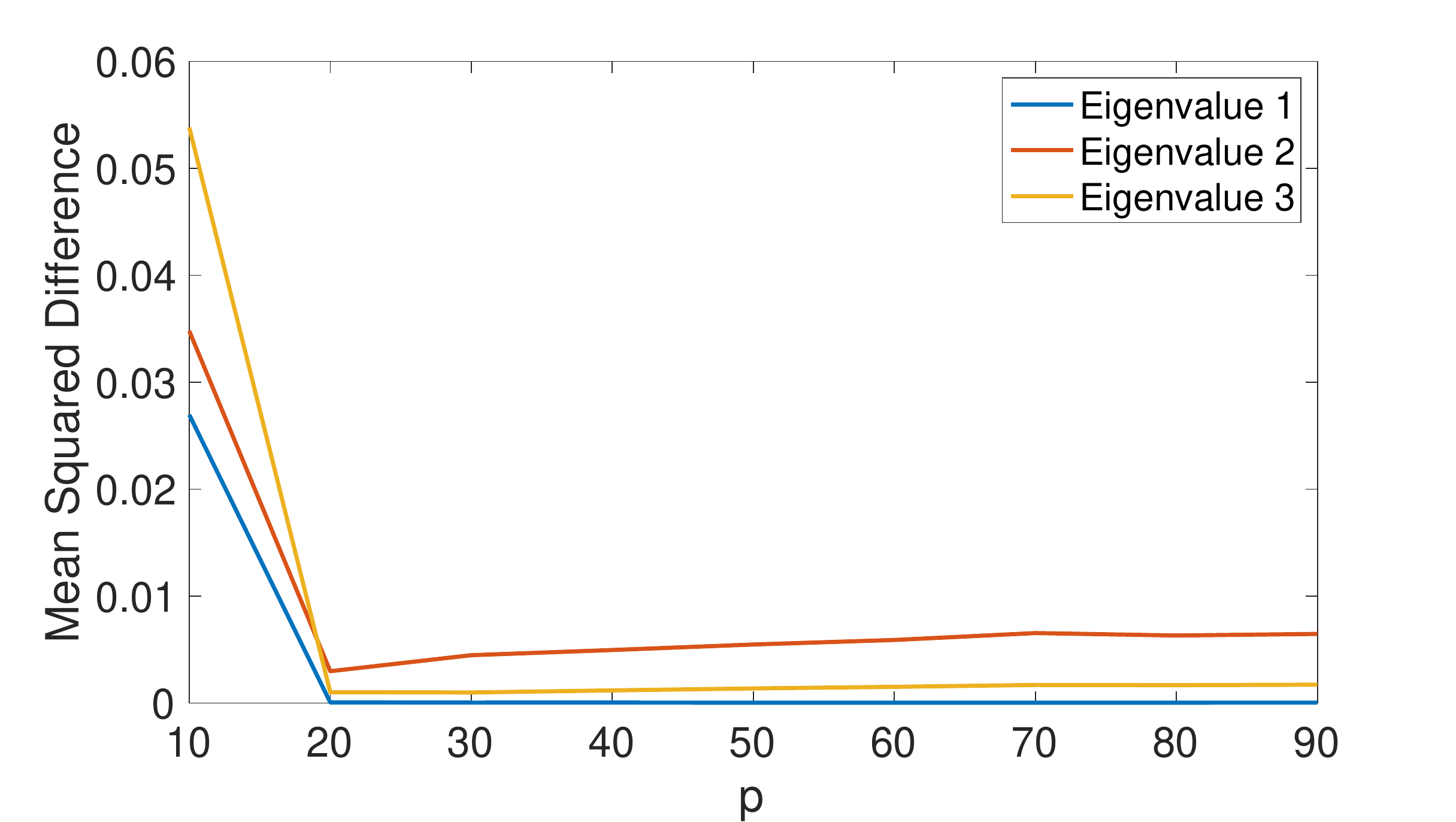}
    \caption{Evaluation of the bagging sub-selection of data with medium noise.  In this case, there are $m=100$ data snapshots and $p$ randomly sub-selected snapshots for bagging.  The mean squared difference between the true value and mean across trials for various $p$ are shown.  Note that only $p=20$ sub-selected snapshots are required to achieve convergence.}
    \label{fig:pareto}
\end{figure}

\section{A Simple Example}
We first tested the performance of the BOP-DMD algorithm on a simple example with additive noise. Algorithm performances were assessed by comparing the computed eigenvalues using optDMD and BOP-DMD to the analytical eigenvalues, as highlighted in previous works~\cite{dawson2016ef,hemati2017tcfd,askham2017arxiv,azencot2019consistent}.
\subsection{Generation of Data}
The data used in this example was generated using the following equation:
\begin{equation}
f(x,t) = g_1(x)e^{\lambda_1 t} + g_2(x)e^{\lambda_2 t} + g_3(x)e^{\lambda_3 t} + \sigma\epsilon\\
\label{eq:toy_example}
\end{equation}
where $\epsilon \sim N(0,1)$ is normally distributed noise and
\begin{align*}
g_1(x) &= \sin(x)\\
g_2(x) &= \cos(x)\\
g_3(x) &= \tanh(x).
\end{align*}
Here, $\sigma$ ranges from 0.001 to 0.05 (depending on noise setting), $x \in [0,1]$, and $t \in [0,1]$. To create the data matrix $X \in \mathbb{R}^{m x n}$, Equation \ref{eq:toy_example} was evaluated on a gridspace comprised of equally spaced spatial points, $x_k$, for $k = 1...n$, and equally spaced time points, $t_k$, for $k = 1...m$. These sampling points are used to construct an individual state space estimate ${\bf x}_k={\bf x}(t_k)$ and the full data matrix $\bX\in\mathbb{C}^{n\times m}$.

\subsection{Methods and Results}
In order to create an initial seed for the BOP-DMD algorithm, the original optDMD algorithm was run on the full data matrix, $\bX$, without initial conditions. The eigenvalues outputted by this algorithm, $\bOmega_{opt}$, were used as initial conditions for our algorithm. For $k$ trials, eigenvalues of various subsets were numerically computed using the aforementioned initial conditions. As shown in Figure \ref{fig:BOPDMD}, these subsets, $\bX^{(k)}\in \mathbb{R}^{nxp}$, for $k = 1...100$ , were generated by concatenating $p$ randomly selected column indices of the original data matrix, $\bX$.

Figure \ref{fig:noise_cycles} shows the BOP-DMD distributions of the computed eigenvalues for each of these subsets. These distributions were generated for various noise realizations (a-c). In each of the distributions, the mean computed eigenvalue using bagging performed better than the original optDMD algorithm. More importantly, this trend in performance is present in each of the three noise realizations, highlighting the robustness and efficacy of this algorithm.


Figure \ref{fig:pareto} shows the squared difference in error between the mean computed eigenvalue and true eigenvalue for medium noise setting ($\sigma = 0.005$) for various batch sizes. As the batch sizes, p, increases, this error appears to plateau. Out of m = 100 snapshots, only 20 randomly selected snapshots are required for convergence. More importantly, optimal performance is actually achieved with p = 20 for a medium noise setting.


\begin{figure*}[t]
        \vspace*{-.8in}
    \centering
    \hspace*{-.93in}
        \begin{overpic}[width =1.3\textwidth]{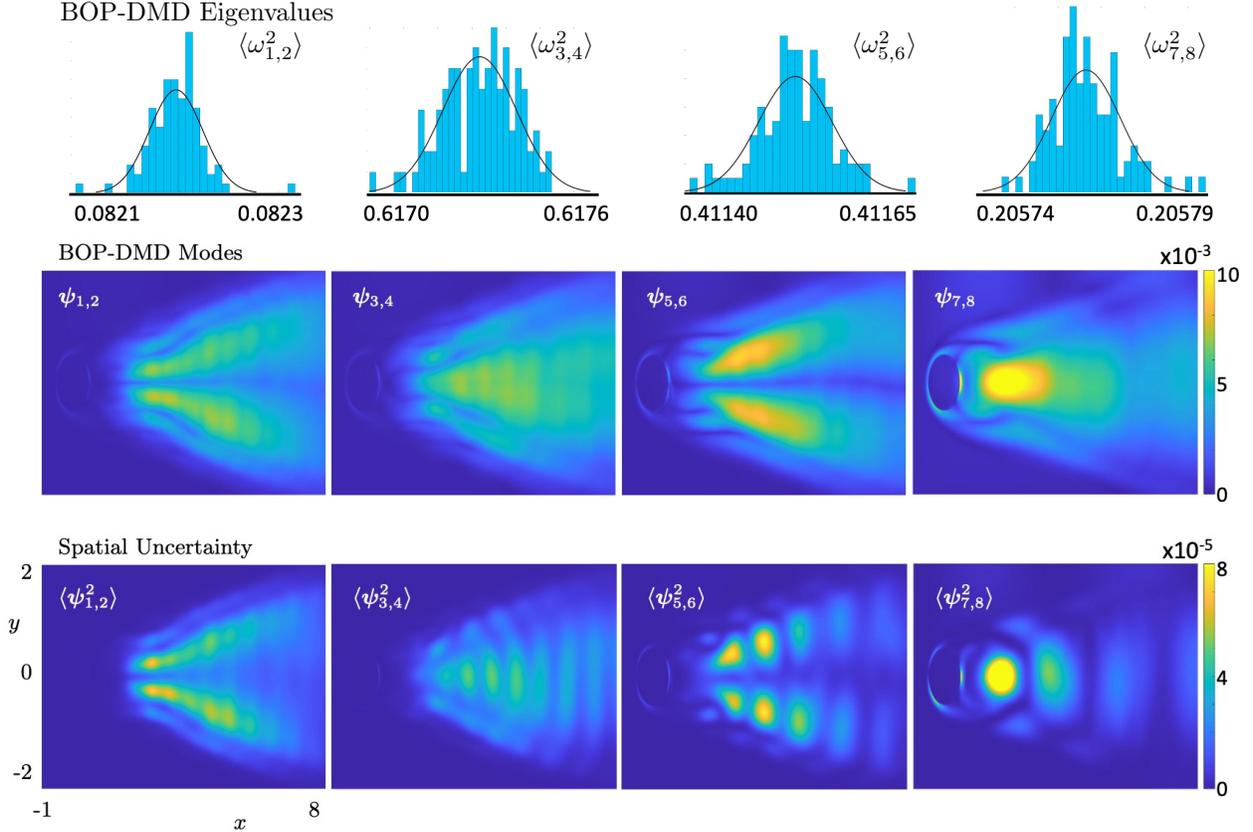}
        \put(25,60){$\langle \omega^2_{1,2} \rangle$}
        \put(43,60){$\langle \omega^2_{3,4} \rangle$}
        \put(63,60){$\langle \omega^2_{5,6} \rangle$}
        \put(81,60){$\langle \omega^2_{7,8} \rangle$}
        \put(14,62){BOP-DMD Eigenvalues}
        \end{overpic}
    \vspace*{-1.3in}
    \caption{Vortex Shedding Example with noise and corruption (first 8 paired modes): Top panel: Temporal uncertainty quantification for eigenvalues corresponding to modes.  The black lines represent a least-square fit of a normal distribution.  Middle panel: Mean of the BOP-DMD modes generated after bagging with $K=100$ trials. Bottom panel: Spatial variance of the uncertainty over the $K=100$ ensemble .}
    \label{fig:fluids_bop2}
\end{figure*}

\begin{figure*}[t]
\vspace*{-.5in}
    \centering
            \begin{overpic}[width =1.0\textwidth]{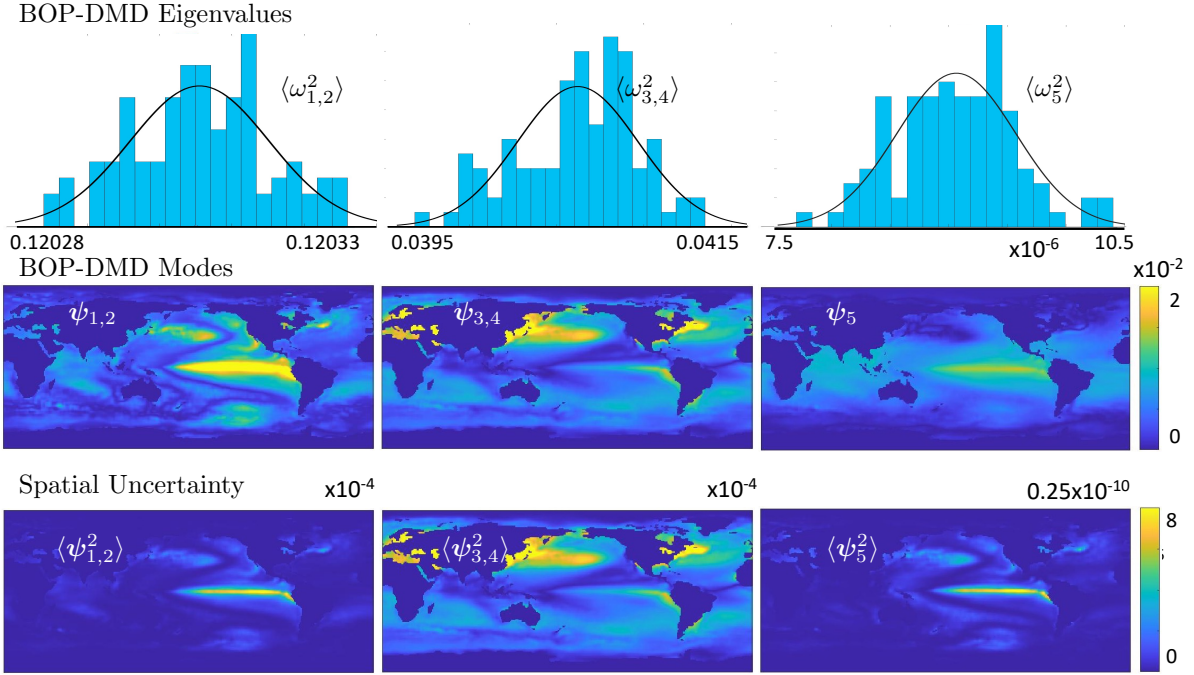}
        \put(25,60){$\langle \omega^2_{1,2} \rangle$}
        \put(52,60){$\langle \omega^2_{3,4} \rangle$}
        \put(85,60){$\langle \omega^2_{5} \rangle$}
        \put(4,66){BOP-DMD Eigenvalues}
       \put(7,23){{\color{white}{$\langle \boldsymbol{\psi}^2_{1,2} \rangle$}}}
       \put(38,23){{\color{white}{$\langle \boldsymbol{\psi}^2_{3,4} \rangle$}}}
       \put(69,23){{\color{white}{$\langle \boldsymbol{\psi}^2_{5} \rangle$}}}
       \put(8,42){{\color{white}{$ \boldsymbol{\psi}_{1,2}$}}}
       \put(39,42){{\color{white}{$ \boldsymbol{\psi}_{3,4} $}}}
       \put(69,42){{\color{white}{$ \boldsymbol{\psi}_{5} $}}}
       \put(4,45.5){BOP-DMD Modes}
       \put(4,28){Spatial Uncertainty}

        \end{overpic}
\vspace*{-1.1in}
%


    \caption{SST data (first 6 modes): Top panel: Uncertainty quantification for eigenvalues corresponding to modes.  The black line represents a least-square fit of a normal distribution.  Middle panel: Mean of the DMD modes generated after bagging with $K=100$ trials. Bottom panel: Spatial variance of the BOP-DMD modes over the ensemble of models.}
    \label{fig:sst_bop}
\end{figure*}



\begin{figure*}[t]
\vspace*{-.0in}
    \centering
%
 \begin{overpic}[width =1\textwidth]{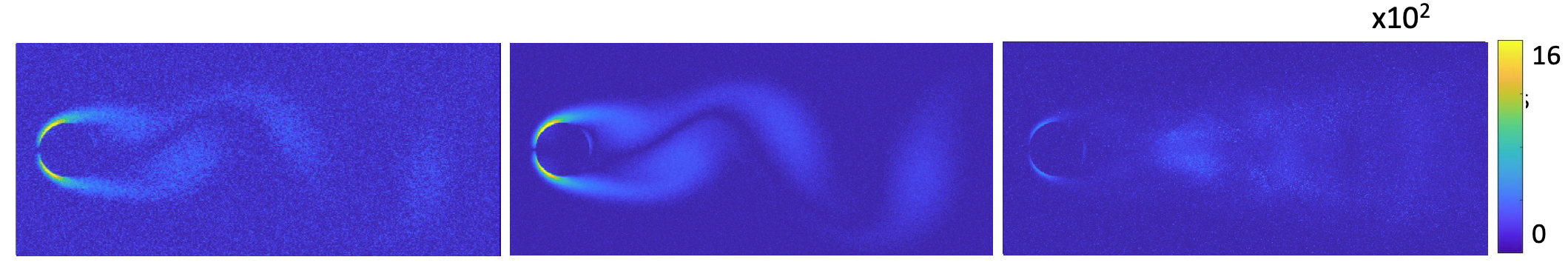}
\end{overpic}
\vspace*{-1.8in}
    \caption{Forecasting skill of BOP-DMD for flow around cylinder with noise as illustrated in Fig.~\ref{fig:fluids_bop2}.  The BOP-DMD model is trained over $t\in[0,100]$ and forecast in the time range $t\in[100,150]$.  For this data set, the BOP-DMD model provides an accurate and stable prediction with nearly vanishing variance. The top panel shows the average vorticity data (black line) as a function of time along with the BOP-DMD reconstruction (red line) and forecast (magenta line).  The slightly visible cyan line in the forecast shows the variance around the average of the magenta forecast line.}
    \label{fig:forecast_flow}
\end{figure*}

\begin{figure*}[t]
\vspace*{-.0in}
    \centering
%
 \begin{overpic}[width =1\textwidth]{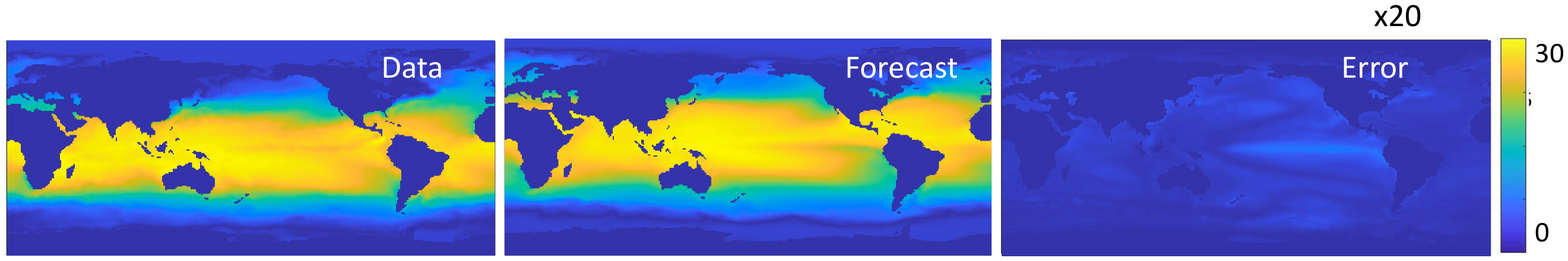}
\end{overpic}
\vspace*{-3.2in}
    \caption{Forecasting skill of BOP-DMD for global sea-surface temperature.  The BOP-DMD model is trained over a 19.2 year period (1000 snapshots) and forecast to the next 7.6 years (400 snapshots).  For this data set, the BOP-DMD model provides an accurate and stable prediction with nearly vanishing variance with five modes.  The error in the right panel is multiplied by 20 for visibility on the same color scale. }
    \label{fig:forecast_flow2}
\end{figure*}

\section{BOP-DMD Applications}

In what follows, we apply the BOP-DMD algorithm to two spatio-temporal systems: flow around a cylinder and sea-surface temperature data.  Both provide a platform for the evaluation of the algorithm's performance, including its ability to produce UQ metrics in both space and time.

\subsection{Flow Around a cylinder}

As a common DMD example~\cite{bagheri2014}, we simulate the fluid flow past a circular cylinder with the two-dimensional, incompressible Navier-Stokes equations at $Re=100$: 
\begin{equation}
\nabla \cdot \boldsymbol{u} =0, \,\,\,\, 
\partial_{t} \boldsymbol{u}+(\bd{u} \cdot \nabla)\bd{u}=-\nabla p+\frac{1}{R e} \Delta \boldsymbol{u}
\label{eq:navierstokes}
\end{equation}
where $\boldsymbol{u}$ is the two-component flow velocity field in 2D and $\boldsymbol{p}$ is the pressure term. For Reynold's number $Re = Re_c \approx 47$, the fluid flow past a cylinder undergoes a supercritical Hopf bifurcation, where the steady flow for $Re<Re_c$ transitions to unsteady vortex shedding~\cite{Bearman69}. The unfolding gives the celebrated Stuart-Landau ODE, which is essentially the Hopf normal form in complex coordinates. This has resulted in accurate and efficient reduced-order models for this system~\cite{Noack2003jfm,Noack2011book}.

Figure~\ref{fig:fluids_bop2} shows the first eight normalized BOP-DMD modes constructed from $K=100$ trials using noisy data.  The modes are cleanly produced and the spatial UQ, as illustrated in the bottom panel of the figure, show the spatial patterns where DMD is most uncertain in its ability to provide an accurate reconstruction.    With $K=100$ ensembles, BOP-DMD is able to both produce UQ metrics for the DMD eigenvalues and their temporal uncertainty as well as a spatial uncertainty map for the variance of each of the modes.  The spatial and temporal UQ provide a framework for Monte Carlo forecasting since the probability distributions of both the spatial and temporal fields can be drawn from to estimate an ensemble of future state predictions.

Although flow around a cylinder is a well known example, it helps illustrates the key concept that BOP-DMD stabilizes the prediction of the DMD modes and eigenvalues by ensembling them.  More than that, the UQ can be explicitly computed for all spatial points and every DMD frequency extracted.  This is in addition to removing DMD bias induced by noise which leads to issues as illustrated in Fig.~\ref{fig:atmos_chem}.

\subsection{Sea-surface temperature}

Another common example used in the DMD and POD literature is sea-surface temperature data.   We consider the NOAA$\_$OISST$\_$V2 global ocean surface temperature data set spanning 1990–2016. The data are publicly available online~\cite{sst}.  The first 1000 snapshots, which ends in 2007 (19.2 years) are used for producing BOP-DMD modes whose resolution is 360x180 pixels.
In the next section, these BOP-DMD modes are used for forecasting.

Figure~\ref{fig:sst_bop} shows the first five BOP-DMD modes along with their DMD eigenvalues.  For these results the
normalized BOP-DMD modes are constructed from $K=100$ of the data.  The modes are cleanly produced and the spatial UQ, as illustrated in the bottom panel of the figure, show the spatial patterns where DMD is most uncertain in its ability to provide an accurate reconstruction.  Note the highest spatial uncertainty along the El Ni\~no Southern Oscillation (ENSO).  The BOP-DMD algorithm also quantifies the temporal uncertainty by allowing for a Gaussian fit to the eigenvalue distribution.

\section{Forecasting with BOP-DMD}

The BOP-DMD algorithm can produce an approximation to the solution (\ref{eq:dmd_opt}).  This solution gives the capability to forecast future states of temporal systems, including the examples considered here of flow around a cylinder and sea-surface temperatures.  Moreover, by using the BOP-DMD forecasting algorithm, Monte Carlo simulations can produce UQ for the future state by drawing from the spatial and temporal distributions of the DMD modes and eigenvalues as shown in Fig.~\ref{fig:fluids_bop2} and \ref{fig:sst_bop}.

The forecasting capabilities for the two examples shown in the previous section are shown in Fig.~\ref{fig:forecast_flow} and \ref{fig:forecast_flow2}.  For the flow around a cylinder, the average vorticity over the domain is illustrated in the top panel.  The forecast in this data set has very little uncertainty as the dynamics is quite low-dimensional and BOP-DMD has no problem extracting the future state.  For the sea-surface temperature data, The first 19.2 years (1000 snapshots) are used to predict the next 7.6 years (400 snapshots).  The 7.6 year forecast is illustrated along with the error.  Note that the largest error (Fig.~\ref{fig:forecast_flow2}) occurs where the spatial UQ metric is largest as shown in Fig.~\ref{fig:sst_bop}.  Only five modes are used to forecast the sea-surface temperature data.

\section{Conclusions}

In conclusion, the {\em dynamic mode decomposition} has emerged as a flexible, adaptive, robust, and general purpose data-driven method for the analysis and characterization of a broad range of scientific applications~\cite{kutz2016book}.  The regression to a linear dynamical system also provides an interpretable, data-driven framework for modal decompositions whose temporal dynamics are exponentials.  The sheer multitude of variants of the DMD regression framework highlights the broad interest from the community in making the method more stable, robust, and predictive.  The BOP-DMD algorithm outlined here integrates and leverages many advantageous mathematical techniques, including variable projection for the optimization and statistical bagging for creating an ensemble of models. When combined, these two widely used techniques are able to construct a model which is less prone to over-fitting and reduces model variance.  Indeed, the ensemble of BOP-DMD provides both spatial and temporal UQ metrics for reconstruction and forecasting that are currently unavailable in other methods.

Demonstrated in this manuscript is the BOP-DMD algorithm on a number of applications.  The algorithm not only stabilizes the DMD method and consistently removes noise bias, but also provides both a spatial and temporal uncertainty quantification of the linear model.  Thus it improves on the performance of optimized DMD while providing additional and critical insights into the dynamics and its spatio-temporal variability.  The method is adaptive, requires very little data, and provides the capability to produce stable, probabilistic forecasts. Moreover, it does this within the framework of an interpretable modal decomposition that reveals dominant spatio-temporal features in a given data set.

As data-driven methods continue to be developed across the engineering, physical and biological sciences, it is imperative that modeling strategies be capable of handling real data that is often noisy and high-dimensional.  With its ease of use, minimal data requirements and interpretability, DMD provides an exceptional practical alternative to data-driven techniques, such as neural network models.  The ensembling aspect of BOP-DMD also provides a Bayesian perspective on the linear model fitting by providing estimates of the probability distributions for the DMD eigenvalues and DMD eigenfunctions, both of which allow for estimates of the method's forecasting horizon and capabilities.  The open source code provided allows for reproducible and broad usage across the sciences.  Moreover, the underlying algorithm can be used with established DMD pre-processing innovations, including time-delay embeddings, data centering and multi-resolution analysis.\\

\noindent{\bf Code Availability:}
The code used in this paper can be found at https://github.com/dsashid/BOP-DMD. Use of this code requires installation of the initial optDMD framework which can be found at https://github.com/duqbo/optdmd.

\section*{Acknowledgments}
JNK would like to acknowledge insightful conversations with both J. C. Louiseau and S. L. Brunton on stabilizing DMD.
The authors acknowledge sponsorship under the Air Force Office of Scientific Research (AFOSR) grant FA 9550-18-1-9-007. JNK acknowledges support from AFOSR grant FA9550-17-1-0329.

\bibliographystyle{unsrt}
\bibliography{merged}

\end{document}